%% file: main.tex
\documentclass[10pt, conference, letterpaper]{IEEEtran}
\IEEEoverridecommandlockouts
\input{macro_packages_ver10.tex}
\usepackage{cite}
\usepackage{etoolbox}

\def\BibTeX{{\rm B\kern-.05em{\sc i\kern-.025em b}\kern-.08em
    T\kern-.1667em\lower.7ex\hbox{E}\kern-.125emX}}
\begin{document}

\title{\LARGE Scenario-Agnostic Zero-Trust Defense with Explainable Threshold Policy: A Meta-Learning Approach \vskip -5mm}

\author{Yunfei Ge$^*$, Tao Li$^*$, and Quanyan Zhu}
\makeatletter
\patchcmd{\@maketitle}
  {\addvspace{0.5\baselineskip}\egroup}
  {\addvspace{-0.85\baselineskip}\egroup}
  {}
  {}
\makeatother
\maketitle
\def\thefootnote{}\footnotetext{ The authors are with Department of Electrical and Computer Engineering, New York University, NY; E-mail: \{yg2047,taoli,qz494\}@nyu.edu. $*$Yunfei Ge and Tao Li contributed equally to this work.}\def\thefootnote{\arabic{footnote}}
\begin{abstract}

The increasing connectivity and intricate remote access environment have made traditional perimeter-based network defense vulnerable. Zero trust becomes a promising approach to provide defense policies based on agent-centric trust evaluation. However, the limited observations of the agent's trace bring information asymmetry in the decision-making. To facilitate the human understanding of the policy and the technology adoption, one needs to create a zero-trust defense that is explainable to humans and adaptable to different attack scenarios. To this end, we propose a scenario-agnostic zero-trust defense based on Partially Observable Markov Decision Processes (POMDP) and first-order Meta-Learning using only a handful of sample scenarios. The framework leads to an explainable and generalizable trust-threshold defense policy. To address the distribution shift between empirical security datasets and reality, we extend the model to a robust zero-trust defense minimizing the worst-case loss. We use case studies and real-world attacks to corroborate the results.

\end{abstract}

\begin{IEEEkeywords}
Zero-trust security, meta learning, scenario-agnostic, threshold policy 
\end{IEEEkeywords}
\section{Introduction}
The recent advances in cloud services, data communication, and automation technologies have increased flexibility and efficiency in modern network systems \cite{zhu2020cross}. However, the adoption of smart devices and the Internet of Things (IoT) has brought up new and expanding cyber risks, not just capable of impacting a particular device but creating severe concerns in the whole system \cite{zhu2021cybersecurity}. The increasing connectivity, heterogeneity, and dynamic accessing environments inevitably enlarge the attack surface and lead to multiple vulnerabilities that attackers can exploit. In response to the vulnerabilities in traditional perimeter-based network security, modern networks must transform from static and perimeter-based defenses to a zero-trust security framework that forfeits the assumption that everything behind the security perimeter is safe. It eliminates implicit trust in each agent and makes defense decisions based on continuous trust evaluation \cite{rose2020zero}.

However, designing the zero-trust defense is not trivial,  and several challenges arise. First, the limited observations of the agent's trace bring information asymmetry in the decision-making. Hence, a quantitative metric measuring the agents' trustworthiness using partial observations is indispensable. Moreover, the zero-trust policy must be generalizable to a family of scenarios. Besides, it is necessary to create a zero-trust defense that is explainable to human operators who develop security solutions based on the reasoning of the defense policy.  In this way, explainable and adaptable zero-trust defenses make the configuration of the large-scale network easier and facilitate the broad adoption of the technology. 

To equip the zero-trust defense  with adaptability under information asymmetry, we propose a new zero-trust defense taking a threshold form based on the concept of meta-learning \cite{meta_survey}. Unlike classical learning paradigms (i.e., reinforcement learning), meta-learning aims to learn a learning strategy using previous training data, rather than a simple decision-making model. In the face of a new task unseen in the training phase, the obtained learning strategy enables the defender to learn a satisfying policy using far less data than from scratch.

Taking inspiration from meta-learning, the proposed ZTD, referred to as scenario-agnostic zero-trust defense (SA-ZTD), enables the defender to adapt to new scenarios (system configurations/attack capabilities) with a modest amount of partial observations using a  learning strategy learned from the training experience. To be more specific, we first formulate a zero-trust network security model using parameterized Partially Observed Markov Decision Processes (POMDP) \cite{zhu2020introduction}, where the parameters represent attack scenarios with distinct system vulnerabilities and the attacker's capabilities. 

\begin{figure}[t!]
\abovecaptionskip=1pt
\centerline{\includegraphics[width=0.9\linewidth]{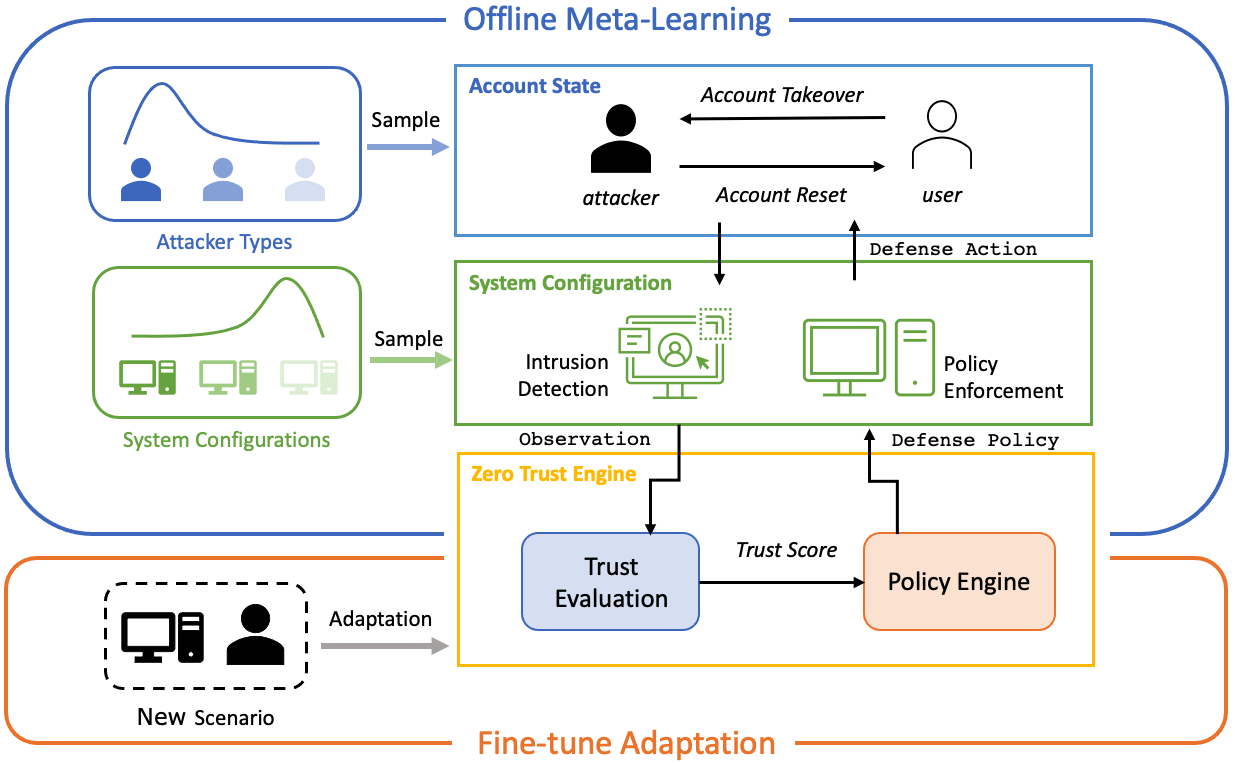}}
  \caption{Illustration of the scenario-agnostic zero-trust defense. The meta policy acquired in the meta-learning phase can respond to unknown attack scenarios with fine-tuned adaptation. }
  \vskip -5mm
  \label{fig:general}
\end{figure}

Since real-world applications involve a large (possibly infinite) number of attack scenarios, it is intractable to compute the optimal policy for each scenario. The proposed SA-ZTD resolves this issue by learning a meta policy and an adaptation mapping (i.e., the learning strategy) using a handful of known scenarios. When deployed in a new attack scenario, the learned mapping can quickly adapt the meta policy to the current environment using a few partial observations. We use the word “agnostic” (whose root means “not known”) to emphasize that the adaptation ability is acquired without knowledge of every scenario. An illustration of SA-ZTD is presented in \Cref{fig:general}. 
 \textbf{Our contributions} are as follows. 1) We propose scenario-agnostic zero-trust defense, a generalizable defense strategy that does require access to every scenario. In addition, SA-ZTD leads to explainable trust-threshold policies. 2) To address the distribution shift between empirical security datasets and reality, we propose a robust zero-trust defense based on SA-ZTD that minimizes worst-case loss. 3) A first-order meta-learning algorithm is developed to learn SA-ZTD efficiently using only a handful of sample scenarios. 

\textbf{Related Works}:
Zero-trust defense has become an emerging trend for addressing the challenges in modern network security. A conceptual zero trust strategy is proposed in  \cite{mehraj2020establishing}  to establish the trust notion in cloud computing.  \cite{ramezanpour2022intelligent} investigates a zero-trust architecture for 5G networks utilizing artificial intelligence to provide information security.
Authors in \cite{dhar2021securing} have combined zero trust and block-chain to manage IoT device security.
Despite the extensive applications, the aforementioned frameworks are conceptual, and the proposed zero-trust defenses highly depend on the underlying system configuration. There is a lack of adaptation/generalization ability in their defense strategy design, and our work is among the first endeavor to investigate adaptable zero-trust defense leveraging meta-learning.

\section{Zero-Trust Defense under Asymmetric Information}
% One important pillar of zero-trust defense is agent-centric
% trust evaluation, where each agent's trust score is updated according to its footprint. However, asymmetric information structures in modern network systems (the defender and the attacker possess different information) cause uncertainty in the trust evaluation and following defense decision-making.  

% Many security problems share asymmetric information structures, in particular, the account
% takeover attacks (ATA) in network systems considered in this work. The following subsection formulates the zero-trust defense problem in ATA as a Partially Observed Markov Decision Process (POMDP).  

Consider account take-over attacks (ATA), where the attacker attempts to compromise the legitimate account using social engineering \cite{heartfield2015taxonomy} and zero-day vulnerabilities \cite{ablon2017zero} at every time step. Meanwhile, the system administrator (the defender) can turn the adversarial account into a legitimate one by removing the stored credentials and resetting the account. 

The true nature of the account, while revealed to the attacker, remains hidden from the defender: only the attacker knows whether the attack succeeds or not. In contrast, the defender can only observe the footprints of the agent through system alerts and monitoring information, such as intrusion-detection systems. Note that the system alerts do not equate to the actual state of the account, as the monitoring mechanism may produce false positive/negative results. 

The asymmetric information structure in ATA, as in other security problems \cite{tao_info}, complicates the defender's decision-making process. The defender needs to dynamically evaluate the trust using partial observations and then decide whether disable the account at the policy enforcement point. The above zero-trust defense problem can be formulated as a POMDP defined by the tuple parameterized by $\theta\in \Theta$: $\left\langle \mathcal{S},\mathcal{A},\mathcal{O}, T_\theta, O, C ,\rho \right\rangle$ discussed in the following.
%, and the following presents each component of the tuple.  
$k\in \{0,1,\dots\}$ denotes the discrete time index.
\begin{itemize}
    \item $\Theta$ denotes the set of all attack \textbf{scenarios}, and each parameter $\theta\in \Theta$ captures the system vulnerabilities and attacker capabilities that affect the system transition to be defined later. 
    \item $\mathcal{S}=\{0,1\}$ denotes the set of  \textbf{account states}. $s=0$ stands for the adversarial, while $s=1$ for the legitimate;
    \item $\mathcal{A}=\{0,1\}$ is the set of \textbf{defense actions}. $a=0$ means removing the account credentials and resetting the account.  $a = 1$ indicates that the defender takes no action;
    \item $\mathcal{O}=\{0, 1\}$ denotes the set of \textbf{system observations}. $o=0$ indicates the defender observes a security alert about the anomaly behaviors of the account, and no alert is observed when $0=1$;
    \item $T^\theta(a)\in \R^{|\mathcal{S}|\times|\mathcal{S}|}$, for $a\in \mathcal{A}$ is the \textbf{system transition matrix} under action $a$. The $ij$-entry of the matrix given  indicates the probability that the account of state $i$ being changed to state $j$, i.e.,  $T^\theta_{ij}(a)=\pr\left(s^{k+1}=j|s^k=i,a^k=a, \theta\right), i,j\in \mathcal{S}.$
    
    Since the defense action is binary, we divide the presentation of the transition matrices into two cases below. 1) \textbf{Active Defense Case}: $a=0$. The defender decides to reset the account, and as a result, the attacker loses its foothold within the system with probability $1-p_a^d$. $p_a^d\in [0, 1]$ is the probability    that the attacker bypasses the defense  when controlling the account. On the other hand, if the account is legitimate, the attacker can compromise it with probability $p_u^d\in [0, 1]$. In summary, $p_a^d$ and $p_u^d$ represent the attacker's capabilities in launching and sustaining ATA, and the transition matrix under active defense is given by $T_{00}(0)=p^d_{a}, T_{01}(0)=1-p^d_{a}, T_{10}(0)=p^d_{u}, T_{11}(0)=1-p^d_{u}$.

 2) \textbf{Normal Operation Case}: $a=1$. In this case, no defense action is taken, and $p_u^n\in [0, 1]$ is the probability that the legitimate account is taken over by the attacker. The magnitude of this value reflects the system vulnerability: the larger $p_u^n$, the more vulnerable the system is. Let $p_a^n\in [0, 1]$ be the probability that the attacker in the system without being detected during normal operation (stealthiness), and the transitions are as below: $T_{00}(1)=p^n_{a}, T_{01}(1)=1-p^n_{a}, T_{10}(1)=p^n_{u}, T_{11}(1)=1-p^n_{u}$.

The system transition matrices are jointly determined by the system vulnerability ($p_u^n$) and the attacker's capability/stealthiness ($p_a^d, p_u^d, p_a^n$). Hence, each attack scenario is fully captured by the concatenation of the relevant parameters in the transition, i.e., $\theta=(p_a^d, p_u^d, p_a^n,p_u^n)\in [0, 1]^4$. 
     
    \item $O\in \R^{|\mathcal{S}|\times|\mathcal{O}|}$ is the observation matrix whose $ij$-entry is defined as $O_{00}=q_{a}, O_{01}=1-q_{a}, O_{10}=q_{u}, O_{11}=1-q_{u}$, where $q_{a}, q_{u}\in[0,1]$ are the detection rate and false alarm rate of the intrusion detection system, respectively.

\item $C(s,a)\in \mathbb{R}$ is the defender's cost when implementing $a$ at state $s$. In particular, the cost against the attacker during normal operation $C(1,0)$ indicates the potential damage of the attack. The defense cost $C(\cdot,1)$ captures the time delay and performance degradation due to the account reset. The defense performance is evaluated by the cumulative cost discounted by the factor $\rho\in(0,1)$. The defender's objective is to find an optimal policy that balances security and system performance across different scenarios to be detailed in the next section.  
\end{itemize}

% Before concluding this section, we present several standing assumptions regarding the ZTD in ATA, which later lead to structural results of the zero
% trust defense policy. 
% \begin{assumption}[Decreasing cost in state variables]
% \label{ass:cost}
% In active defense, the defender spends more effort clearing the
% stored credentials if the account is taken control
% by an attacker: $C(0,0)\geq C(1,0)$. During normal operation, the ATA attack undermines the system performance: $C(0,1)\geq C(1,1)$
% \end{assumption}

% \begin{assumption}[Submodular cost]
%     \label{eq:sub-cost}
%     Switching from normal operation to active defense brings a more significant performance improvement when the account has been taken over: $C(0,1)-C(0,0) \geq C(1,1)-C(1,0)$
% \end{assumption}

% \begin{assumption}[Effective detection]
% The attacker has a higher chance of raising a security alert
% than the user, i.e.,  $q_a \geq q_u$.
% \end{assumption}

% \begin{assumption}
%     For the attacker, holding the compromised account is more likely to succeed than taking over a legitimate one, whatever the defense action is: $p^d_a\geq p^d_u$ and $p^n_a\geq p^n_u$.
% \end{assumption}

% \begin{assumption}
%     Switching from normal operation to active defense $p^n_a-p^d_a \geq p^n_u-p^d_u$.
% \end{assumption}

\section{Scenario-Agnostic Zero-Trust Defense}
For a specific attack scenario $\theta$, a POMDP-based zero-trust defense under asymmetric information is developed in \cite{ge22trust}, where the defender dynamically evaluates the account's trustworthiness using the \textit{trust score (TS)}, defined as the belief that the user is legitimate, i.e., $TS^k:=b^k(s=1)$. The defender's belief is updated in the following Bayesian manner: for $s'\in \mathcal{S}$,
\begin{align*}
    b^{k}(s')=\frac{O_{s^\prime,o^{k+1}}(a^k)\sum_{s\in\mathcal{S}}T_{s,s^\prime}(a^k)b^k(s)}{\sum_{s^\prime\in \mathcal{S}}O_{s^\prime,o^{k+1}}(a^k)\sum_{s\in\mathcal{S}}T_{s,s^\prime}(a^k)b^k(s)}.
\end{align*}
Based on this trust evaluation, the defender searches for a defense policy in  $\Pi:=\{\pi|\pi:\Delta(\mathcal{S})\rightarrow \mathcal{A}\}$ to minimize the expected cumulative cost $ U_\theta(\pi)=\mathbb{E}_{a^k\sim \pi(b^k),T^\theta, O}[\sum_{k=0}^\infty\rho^k c(s^k,a^k)\mid b^0]$, where $b^0\in \Delta(\mathcal{S})$ specifies the initial trust of the account.

A stochastic gradient descent algorithm is proposed in \cite{ge22trust} to minimize the cost $U_\theta(\pi)$, leading to a simple approach  to zero-trust defense (ZTD). However, the resulting optimal policy is generally scenario-dependent: the defense policy is designed for some specific system configuration and  attacker capabilities. The insufficiency of this approach is the lack of \textbf{adaptation/generalization} to different scenarios. As observed in our experiments, a slight change in $\theta$ leads to differences in defense policies %(see Figures \ref{fig:original_sys}, \ref{fig:original_att})
, meaning that the defense policy for one scenario does not generalize to another.  

To equip the ZTD with adaptation/generalization ability, we propose a new zero-trust approach based on the meta-learning idea.  As a learning-to-learn approach, meta learning \cite{meta_survey} intends to build an internal representation of the policy (meta policy) that is broadly suitable for a collection of different but related scenarios. When deployed in a specific scenario possibly unseen in the training phase, the meta policy can quickly adapt to the current environment using a few data, where the adaptation strategy is learned from past training data. In short, two pillars of meta learning are the meta policy $\pi_{meta}$ and the adaptation mapping $\phi:\Pi\times \Theta\rightarrow\Pi$, respectively. Leveraging the above notions, we present a meta-learning-based ZTD in the following. 
\begin{definition}[Scenario-Agnostic Zero-Trust Defense]
   A pair $\left\langle \pi_{meta}, \phi \right\rangle$ is said to be a scenario-agnostic zero-trust defense (SA-ZTD) with respect to a scenario distribution $p\in \Delta(\Theta)$ if the pair solves for the minimization problem below
   \begin{align}
       \min_{\pi,\phi}\E_{\theta\sim p}[U_\theta(\phi(\pi,\theta))]
       \label{eq:meta-obj}
   \end{align}
\end{definition}
\vskip -2mm
Such a defense is scenario-agnostic in that solving for \eqref{eq:meta-obj} (approximately) does not require the knowledge of every scenario $\theta\in \Theta$ nor the scenario distribution $p$. Similar to empirical risk minimization (ERM) \cite{vapnik1999nature}, a solution to \eqref{eq:meta-obj} is obtained by solving the sample average approximation:
\begin{align}
    (\pi_{meta},\phi)\in \argmin \frac{1}{|\widehat{\Theta}|}\sum_{\theta\in \widehat{\Theta}}U_\theta(\phi(\pi,\theta)),
    \label{eq:meta-emp}
\end{align}
\vskip -2mm
where $\widehat{\Theta}\subset\Theta$ is a finite collection of scenarios i.i.d. sampled from $p\in \Delta(\Theta)$. The term ``agnostic'' emphasizes that the exact scenario distribution $p$ is usually unknown in security practice and often replaced by an empirical distribution provided by security datasets, such as the data from MITRE ATT\&CK \cite{strom2018mitre} considered in the experiment section. Using the ERM language, it is expected that the empirical risk minimizer in \eqref{eq:meta-emp} approximates the population risk minimizer in \eqref{eq:meta-obj}.  When dealing with another scenario $\theta$ unseen in the sample set $\widehat{\Theta}$, the adapted policy $\phi(\pi_{meta},\theta)$ achieves satisfying generalization.  

Note that the pair $(\pi_{meta}, \phi)$ in \eqref{eq:meta-emp} aims to minimize the sample average. However, a distribution shift between the empirical one and the true one in \eqref{eq:meta-obj} can reduce the adaptation/generalization ability, as the resulting meta policy may overfit to popular scenarios and perform poorly on rare ones \cite{mehta20curriculum,collins2020task}. To address this issue, one can also design a distribution as presented in \eqref{eq:meta-robust}, leading to a robust ZTD that minimizes the worst-possible loss across all scenarios.  
\begin{definition}[Scenario-Robust Zero-Trust Defense (SR-ZTD)]
A pair  $\left\langle \pi_{meta}, \phi \right\rangle$ is scenario-robust if it solves for 
\begin{align}
    \min_{\pi,\phi}\sup_{p\in \Delta(\Theta)}\E_{\theta\sim p}[U_\theta(\phi(\pi,\theta))].
    \label{eq:meta-robust}
\end{align}   
\end{definition}\vskip -2mm
The empirical approximation to \eqref{eq:meta-robust} using $\widehat{\Theta}$ is given by $\min_{\pi,\phi}\max_{p\in \Delta(\widehat{\Theta})}\E_{\theta\sim p}[U_\theta(\phi(\pi,\theta))]$, which is equivalent to finding the optimum for the worst-possible case: $\min_{\pi,\phi}\max_{\theta\in \widehat{\Theta}}U_\theta(\phi(\pi,\theta))$, since $\Delta(\widehat{\Theta})$ is a probability simplex in a finite-dimensional space, and the support of the worst-case distribution contains some extreme points.

\subsection{Gradient-based Adaptation and Explainable Meta Policy}
To elaborate on  the adaptation mapping in SA-ZTD, we consider the following minimization problem in comparison to \eqref{eq:meta-obj}: $\min_{\pi}\E_{\theta\sim p}[U_\theta(\pi)]$. The corresponding minimizer is denoted by $\pi_{avg}$. When dealing with new tasks that are distant from the majority of training scenarios, $\pi_{avg}$ does not generalize well, as observed in Tables~\ref{tab:adapt_sys} and \ref{tab:adapt_att}.  In contrast, the adaptation $\phi$ in SA-ZTD improves the generalization by updating $\pi_{meta}$ based on a few interactions in the new scenario, leading to  a data-driven adaptation detailed below. 

Since the function class $\{\phi|\phi:\Pi\times \Theta\rightarrow\Pi\}$ is infinite-dimensional, directly seeking an adaptation mapping through \eqref{eq:meta-obj} [or \eqref{eq:meta-emp}] is intractable. One remedy is to restrict the focus to the parameterization class where the mapping is parameterized by  $\gamma\in \R^n$, $n\in \Z_{+}$. For example, $\phi_\gamma$ can be parameterized by recurrent neural networks, where $\gamma$ is the model weights, and the optimal adaptation is determined by  training algorithms \cite{hochreiter01meta-recurrent}.  Another well-accepted parameterization is the gradient-based adaptation: $\phi_\gamma(\pi, \theta):=\pi-\gamma \nabla U_\theta(\pi)$, and $\gamma$ is the gradient step size to be optimized \cite{meta-sgd}. 

We consider the gradient-based adaptation due to its mathematical clarity and computational efficiency \cite{finn17_maml}. The adaptation step size $\gamma$ is assumed constant to simplify the exposition. The objective in SA-ZTD under gradient adaptation is $\min_{\pi}\E_{\theta\sim p}[U_\theta(\pi-\gamma \nabla U_\theta(\pi))]$ (fixing the adaptation), which implies that after obtaining the meta policy obtained through the meta training phase [i.e., solving \eqref{eq:meta-emp}], one-step gradient descent update $\pi_{meta}-\gamma\nabla U_\theta(\pi_{meta})$ suffices for the new scenario when $\pi_{meta}$ is implemented. The gradient $\nabla U_\theta(\pi_{meta})$ in the deployment phase is estimated using a finite-different approximation scheme: simultaneous perturbation stochastic approximation (SPSA) \cite{spsa}, briefly reviewed in Algorithm~\ref{algo}. Such an estimate only requires finite horizon Monte Carlo (MC) simulations within the POMDP, leading to a lightweight adaptation without learning from scratch. 

\paragraph{Meta Threshold Policy} Another reason to focus on gradient-based adaptation is that the resulting meta policy and adapted policy take threshold forms, leading to a switching control. The active defense is implemented when the trust score (TS) is below the threshold. 

To be specific, we first note that for each scenario $\theta$, the optimal policy, under some mild assumptions, is a stationary threshold policy \cite[Theorem 1]{ge22trust}. Hence, it suffices to find the minimizer $\pi^*_\theta$ to $U_\theta(\pi)$ within the set of threshold policies $\Pi$, where each element is parameterized by a threshold $\tau\in[0,1]$: $\pi_\theta(TS)=\mathds{1}_{\{ \tau< TS \leq 1\}}$. Therefore, searching for the optimal policy is equivalent to finding the optimal threshold $\tau^*$. With a slight abuse of notations, we consider the cost $U_\theta(\pi)$ also a function of the threshold $\tau$, i.e., $U_\theta(\pi)=U_\theta(\tau)$.  $U_\theta(\pi)$ and $U_\theta(\tau)$ are used interchangeably when no confusion arises.

To preserve the threshold form, we replace the one-step gradient adaptation with a projected gradient update: $\operatorname{Proj}_{[0,1]}\{\tau-\gamma\nabla U_\theta(\tau)]\}$ so that the updated $\tau$ remains within $[0, 1]$, and the adapted policy is still a valid threshold policy. Hence, the objective in SA-ZTD \eqref{eq:meta-obj} is now given by
\begin{align}
\label{eq:meta-tau}
    \min_{\tau\in [0,1]}\E_{\theta\sim p}[U_\theta(\operatorname{Proj}_{[0,1]}\{\tau-\gamma\nabla U_\theta(\tau)\})]
\end{align}
As a result, the meta policy takes the threshold form that is explainable to human operators, increasing the accessibility and transparency  of learning-based ZTD. However, the price to pay is that \eqref{eq:meta-tau} is a nonsmooth optimization problem, more involved than the original \eqref{eq:meta-obj}. To address this nonsmoothness, we propose an approximate stochastic gradient descent (SGD).   

\paragraph{Meta Learning Algorithm} 
 To solve \eqref{eq:meta-emp} [and \eqref{eq:meta-robust}], consider optimizing the empirical loss through SGD, i.e., $\tau^{t+1}_{meta}=\tau_{meta}^{t}-\frac{\alpha^t}{|\widehat{\Theta}_t|}\sum_{\theta\in \widehat{\Theta}_t} \nabla U_\theta(\phi(\tau^t_{meta},\theta))\nabla_\tau \phi(\tau^t_{meta},\theta)$, where $\alpha^t$ denotes the SGD step size, and $\widehat{\Theta}_t$ is a batch of scenarios sampled from $\widehat{\Theta}$ at $t$-th iteration. In addition to the non-smoothness issue, computing the gradient $\nabla_\tau \phi(\tau^t_{meta},\theta)$ involves a Hessian matrix $\nabla^2 U_\theta$, and the Hessian-vector product computation is costly. To resolve these issues, we ignore term $\nabla_\tau \phi(\tau^t_{meta},\theta)$ as commonly practiced in meta-learning applications \cite{finn17_maml}, and it does not significantly affect the meta-learning performance  as observed in \cite{fallah2020convergence}. 

Finally, it is numerically intractable to compute the exact policy gradient $\nabla U_\theta(\tau)$, and hence, we resort to the SPSA method (reviewed in \Cref{algo}). Denote by $\widehat{\nabla} U_\theta(\tau)$ the SPSA gradient estimate under the policy $\pi$, then the one-step SGD update  can be rewritten as 
\begin{align}
    \tau^{t+1}_{meta}&=\tau_{meta}^{t}-\frac{\alpha^t}{|\widehat{\Theta}_t|}\sum_{\theta\in \widehat{\Theta}_t} \widehat{\nabla} U_\theta(\tau^t_\theta),\label{eq:meta-update}\\
    \tau^t_{\theta}&= \operatorname{Proj}_{[0,1]}\{\tau^t_{meta}-\gamma \widehat{\nabla}U_\theta(\tau^t_{meta})\}.\label{eq:spsa_adapt}
\end{align}
A summary of the above SPSA-based first-order meta-learning algorithm (SPSA-FOML) for SA-ZTD \eqref{eq:meta-obj} is in Algorithm~\ref{algo}.  

As for the minimax problem in SR-ZTD \eqref{eq:meta-robust}, we consider a stochastic gradient descent ascent (SGDA) algorithm, widely employed in adversarial meta learning \cite{collins2020task,tao22sampling}. In SGDA, in addition to performing gradient descent on $\tau^t_{meta}$, a gradient ascent is applied to the variable $p^t\in \Delta(\widehat{\Theta})$: $p^{t+1}(\theta)=p^{t}(\theta)+\beta^t U_\theta(\phi(\tau^t_{meta},\theta))$, for $\theta\in \widehat{\Theta}$, where $\beta^t$ is the ascent step size. Similar to the SGD update in SA-ZTD, the gradient adaptation $\phi(\tau^t_{meta},\theta)$ is performed using SPSA estimate in \eqref{eq:spsa_adapt}. Meanwhile, the value function $U_\theta(\pi)$ is estimated through MC simulation by averaging the cumulative rewards of multiple finite-horizon MC rollouts \cite{Bertsekas1996NeurodynamicP}. Denote by $\widehat{U}_\theta$ the MC estimate, and for all $\theta\in \widehat{\Theta}^t$, the stochastic gradient ascent (SGA) update is given by 
\begin{align}
    p^{t+1}(\theta)=p^{t}(\theta)+\beta^t \widehat{U}_\theta(\tau^t_\theta).
    \label{eq:p-update}
\end{align}
Note that after the SGA update, $p^{t+1}$ shall be projected to $\Delta(\widehat{\Theta})$. As one can see from Algorithm~\ref{algo}, SA-ZTD and SR-ZTD share the same meta-policy update using SGD [the minimization part in \eqref{eq:meta-robust}], and the only difference is that SR-ZTD samples scenarios using the distribution updated by SGA. 
\paragraph{Complexity Analysis}
We briefly discuss the algorithmic complexity using results in \cite{fallah2020convergence} and \cite{tao22sampling}. Note that the bias of the SPSA gradient estimate is of $O[(\eta^t)^2]$ \cite{spsa} (assuming summable). Hence, the summation  $\sum_{t}\E[\|\nabla U_\theta(\tau_{meta}^t)\|^2]$ is controllable \cite[Lemma E.5]{tao22sampling}, which implies that SPSA-FOML converges to the $\epsilon$-first-order stationary point (line 14 in \Cref{algo}) in SA-ZTD \cite[Theorem 5.15]{fallah2020convergence} and SR-ZTD \cite[Theorem E.2]{tao22sampling} within $O(\epsilon^{-2})$ iterations.
\begin{algorithm}
\small
    \caption{SPSA-FOML for SA(SR)-ZTD}
    \label{algo}
    \begin{algorithmic}[2]
        \State \textbf{Initialization}: threshold $\tau_{meta}^0\in [0,1]$, adaptation step size $\gamma$, scenario samples $\widehat{\Theta}$, scenario distribution $p^0=\operatorname{Uniform}(\widehat{\Theta})$, SPSA perturbation sequence $\{\eta^t\}$, gradient descent step sizes $\{\alpha^t\}$, ascent step sizes $\{\beta^t\}$, error tolerance $\epsilon$;
        \For{$t=0, 1,\ldots,$}
        \State Sample a batch of scenarios $\widehat{\Theta}^t$ according to $p^t$;
        \For{every scenario $\theta\in \widehat{\Theta}^t$}
        \State $\widehat{\nabla}U_\theta(\tau_{meta}^t)\leftarrow \verb|SPSA|(\tau_{meta}^t, \eta^t, \theta)$;
        \State $\tau_\theta^t\leftarrow$ gradient-adaptation in \eqref{eq:spsa_adapt};
        \State $\widehat{\nabla} U_\theta(\tau^t_\theta)\leftarrow\verb|SPSA|(\tau_\theta^t, \eta^t, \theta)$;
        \If{SR-ZTD is activated} 
        \State $\widehat{U}_\theta(\tau^t_\theta)\leftarrow$ MC simulation;
        \State $p^{t+1}(\theta)\leftarrow$ one-step SGA in \eqref{eq:p-update}
        \Else 
        \State {$p^{t+1}\leftarrow\operatorname{Uniform}(\widehat{\Theta})$}
        \EndIf
        \EndFor
        \State  $\tau_{meta}^{t+1}\leftarrow$ one-step SGD in \eqref{eq:meta-update}
        \If{$|\widehat{\nabla} U_\theta(\tau_{meta}^t)|\leq \epsilon $ for all $\theta$} break
        \EndIf
        \EndFor
        \State \Return{$\tau^{t}_{meta}$}
        
        \Function{SPSA}{$\tau, \eta, \theta$}\Comment{subroutine for SPSA}
        \State $d\gets \operatorname{Bernoulli}(\{1,-1\},0.5)$;
        \State $\tau_{+}\gets \operatorname{Proj}_{[0,1]}(\tau+\eta d)$; $\tau_{-}\gets \operatorname{Proj}_{[0,1]}(\tau-\eta d)$; 
        \State $\widehat{U}_\theta(\tau_{\pm})\gets$ MC simulation using $\tau_{\pm}$;
        \State $\widehat{\nabla} U_\theta(\tau)\gets \frac{\widehat{U}_\theta(\tau_{+})-\widehat{U}_\theta(\tau_{-})}{2\eta d }$
        \State\Return{$\widehat{\nabla} U_\theta(\tau)$}
        \EndFunction
    \end{algorithmic}
\end{algorithm}
\vspace*{-1em}

\section{Experimental Result}
%We present experimental results to corroborate the adaptability and explainability of SA-ZTD. The experimental evaluation aims to show that the meta-threshold policy is explainable based on scenario distributions and can adapt to the underlying scenario quickly. 
We present experimental results to show that the SA-ZTD policy is explainable based on scenario distributions and can adapt to the underlying scenario quickly. 
For illustrative purposes, we discuss two scenarios:  varying system vulnerability $\theta =(p_u^n)$ and attacker's capability/stealthiness $\theta=(p_a^n)$.

\textbf{Experiment Setup}: The experiment parameters follow the suggestions in \cite{spsa}:  $\eta^t=0.4/t^{0.2}$, $\alpha^t=\beta^t=0.017/(t+50)^{0.602}$, $\gamma = 0.005$, $\epsilon=10^{-3}$. In the MC simulations, we have $\rho=0.86$, horizon length $T = 100$, $|\widehat{\Theta}|=1000$  sampled from distributions to be defined later, and the batch size  $|\widehat{\Theta}^t|=10$. Each evaluation reports the mean and standard deviation under  $50$ repeated runs using different random seeds. 

\vspace*{-0.4em}
\subsection{System Vulnerability}
We consider the following baseline configuration and train the model for different system vulnerability distributions: $p^d_a=0.2,  p^d_u=0.1,  p^n_a=0.8, q_a=0.9,  q_u=0.1, C(0,0)=10,  C(0,1)=15, C(1,0)=3 ,C(1,1)=0$.
To obtain the threshold policy \cite{ge22trust}, we consider the system vulnerability satisfies $0\leq p_u^n\leq \min\{p_a^n,p_a^n-p_a^d+p_u^d\}$, i.e., $p_u^n\in [0,0.7]$. Each $p_u^n$ value represents one scenario $\theta=(p_u^n)$ in our model. The larger $p_u^n$, the more vulnerable the system is. 
\begin{figure}[h!]
\vskip -3mm
\abovecaptionskip=0pt
\centerline{\includegraphics[width=0.83\linewidth]{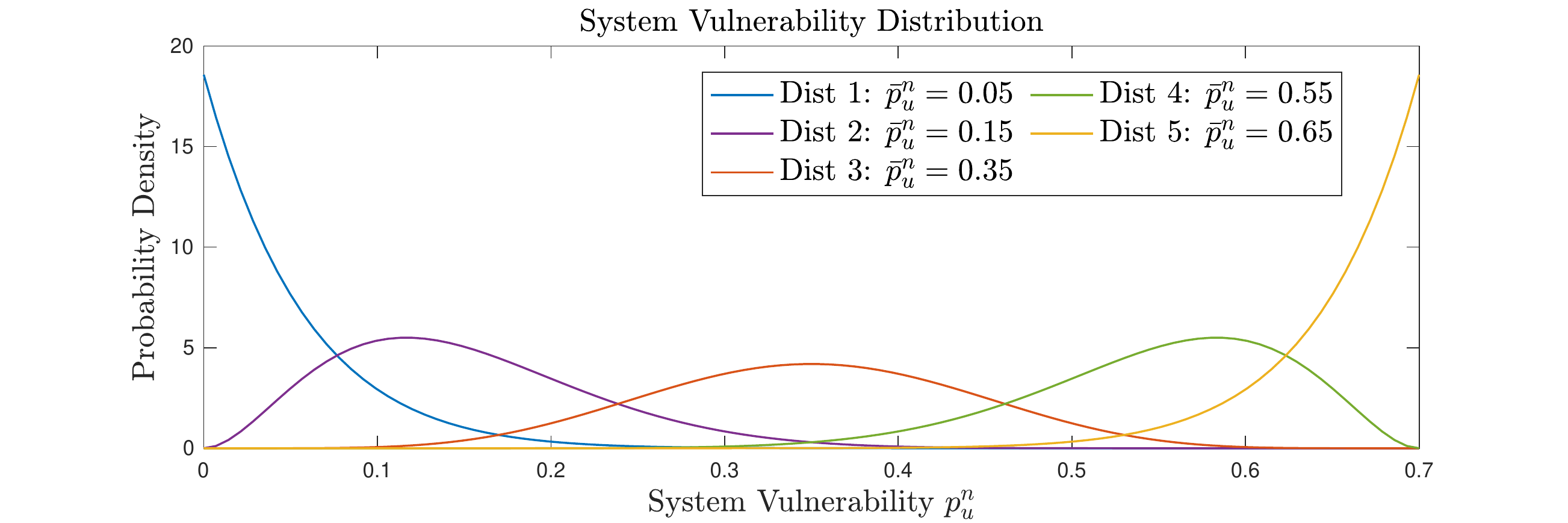}}
  \caption{Illustration of 5 distributions.}
  \vskip -3mm
  \label{fig:dist_pup}
\end{figure}

\subsubsection{\textbf{Explainability}}
%We start with the explainability of the SA-ZTD policy under different system vulnerabilities. 
We consider a generalized Beta distribution family $p_i(\theta,\alpha_i,\beta_i)$ with support $\theta = p_u^n\in [0,0.7]$ to represent the vulnerability distributions.  The $\alpha_i$ and $\beta_i$ values are chosen to generate means $\bar{p}_{u,i}^n=\E_{\theta\sim p_i} [\theta=p_u^n]$, whose values are shown in Figure~\ref{fig:dist_pup}.

In Figure~\ref{fig:pup_policy}, the black line illustrates the optimal policies under each fixed scenario $\theta = p_u^n$. 
We train the meta policy $\tau_{meta}$ under the $5$ distributions and obtain the results, as in Figure~\ref{fig:pup_policy}. The dash lines are the single scenario optimal policies at $p^n_{u,min}$ and  $p^n_{u,max}$.
Experimental results agree with the explanation that if the system on average is more likely to be resistant to attacks (closer to $p^n_{u,min}$), $\tau_{meta}$ is relatively lower and closer to $\tau_{min}$. On the contrary, if the system on average is more vulnerable (closer to $p^n_{u,max}$), the meta policy focuses more on the vulnerable side and increases the minimum acceptable trust score.

\subsubsection{\textbf{Adaptability}}
\begin{figure*}[t!]
          \centering
          \begin{minipage}{.24\linewidth}
            \centering
              {\includegraphics[width=\linewidth]{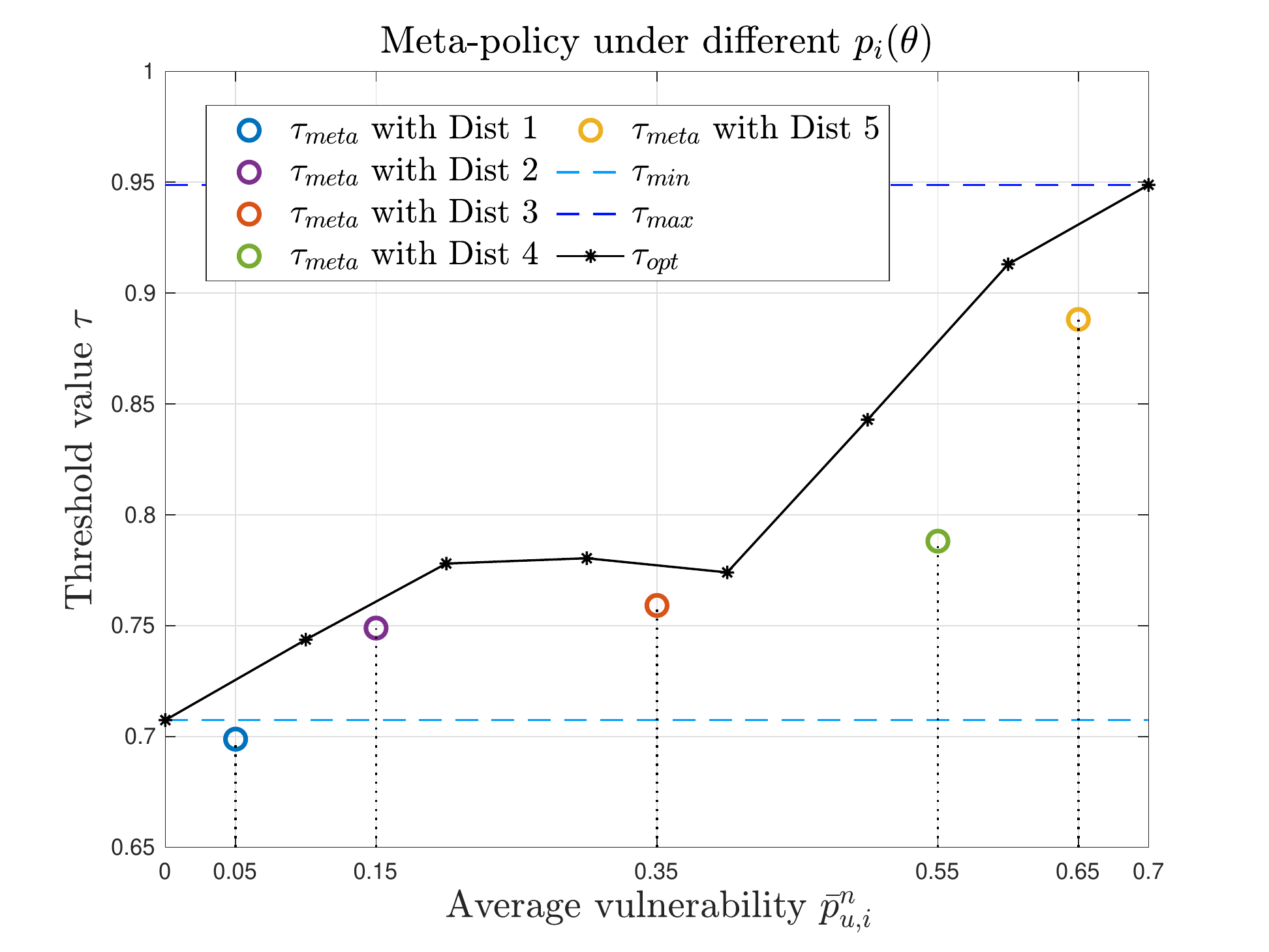}}
              \caption{Meta policy under different system vulnerability distributions.}
          \label{fig:pup_policy}
          \end{minipage}
          \begin{minipage}{.24\linewidth}
            \centering
              {\includegraphics[width=\linewidth]{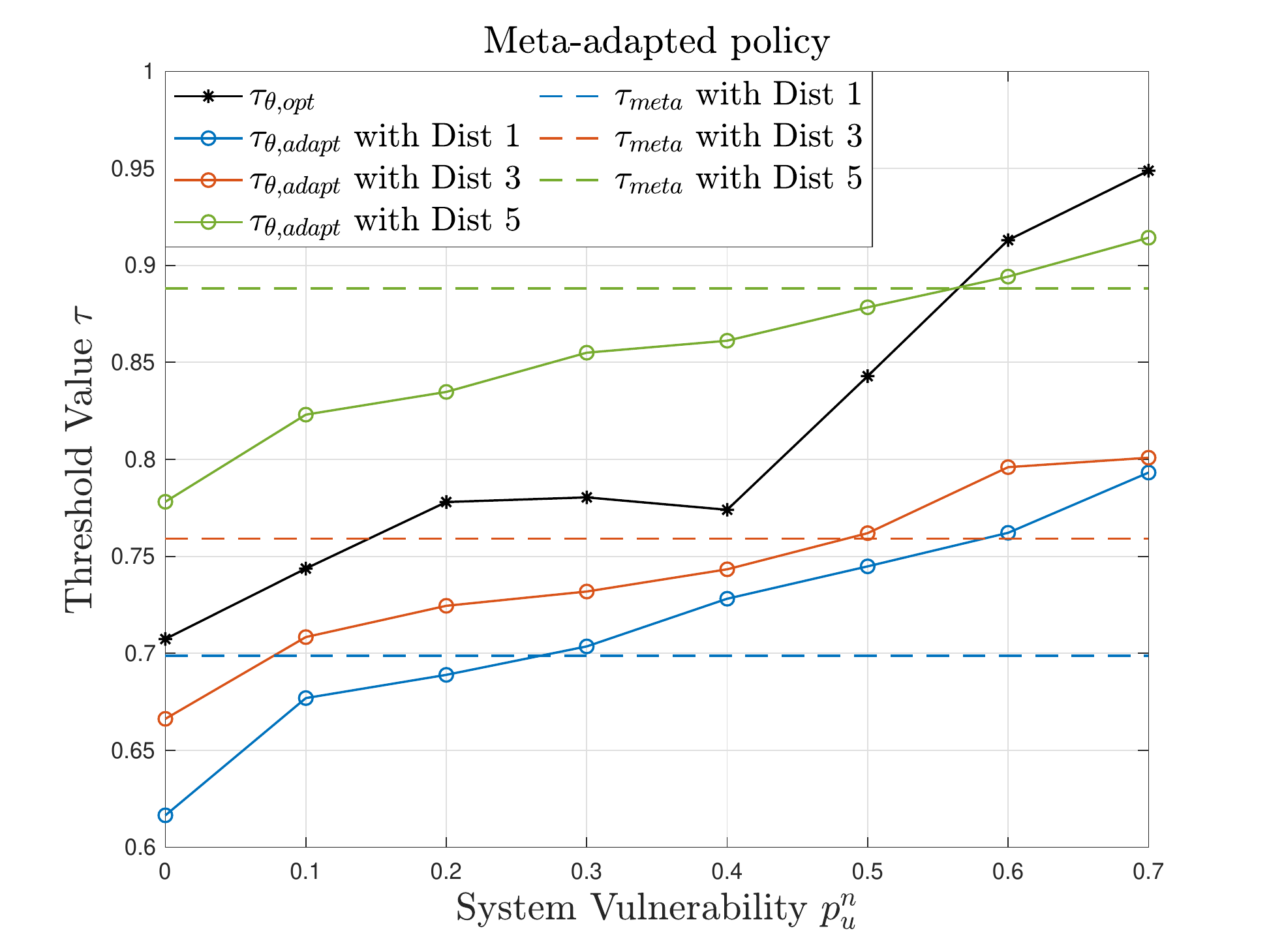}}
          \caption{Adapted policy $\tau_{adapt}$ after one-step gradient update. }
          \label{fig:pup_adapt}
          \end{minipage}
          \begin{minipage}{.24\linewidth}
            \centering
              {\includegraphics[width=\linewidth]{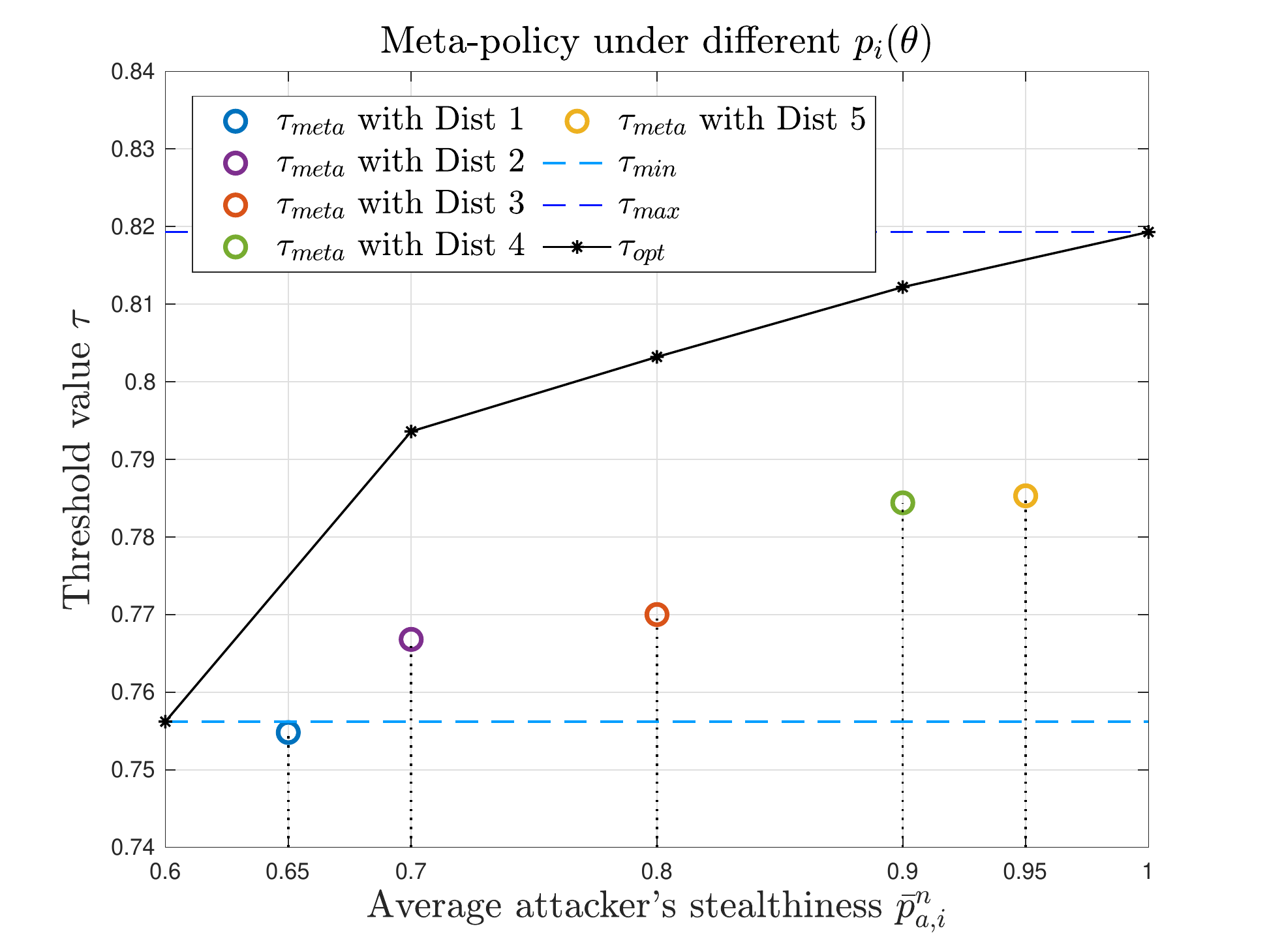}}
          \caption{Meta policy under different attacker stealthiness distributions. }
          \label{fig:pap_policy}
          \end{minipage}
          \begin{minipage}{.24\linewidth}
            \centering
              {\includegraphics[width=\linewidth]{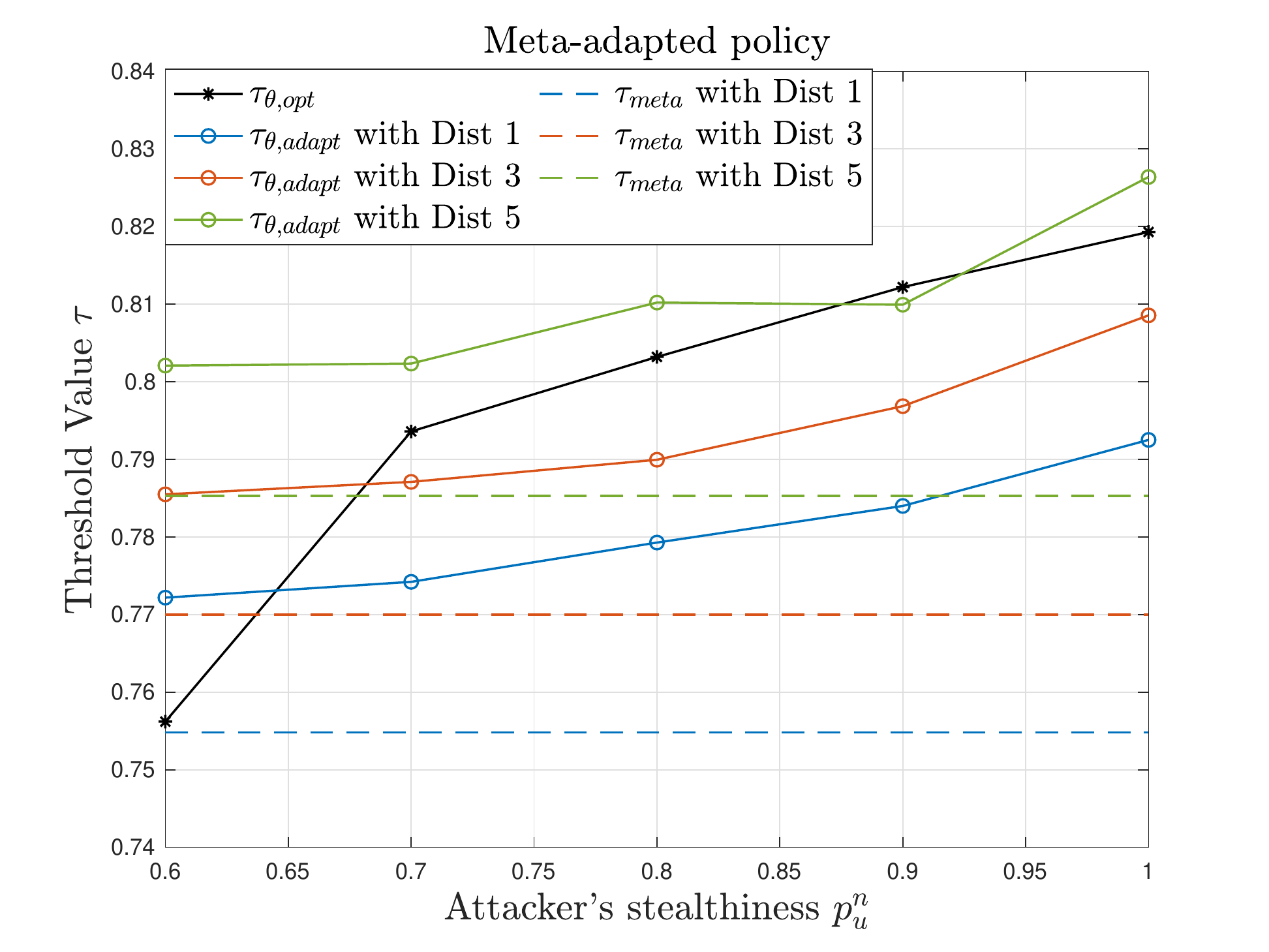}}
          \caption{Adapted policy $\tau_{adapt}$ after one-step gradient update.}
          \label{fig:pap_adapt}
          \end{minipage}
          \end{figure*}

In Figure~\ref{fig:pup_adapt}, we look into the updated policy after a one-step gradient (also called one-shot) adaptation. 
%The black line represents the single scenario optimal policies.
The figure demonstrates that the adapted policies can capture the relationship between the system vulnerability and defense policy.
The $\tau_{\theta,adapt}$ is lower when the actual scenario has lower $\theta=p_u^n$, while $\tau_{\theta,adapt}$  is increased if $\theta=p_u^n$ increases. It indicates that the meta policy can successfully adapt to each scenario. 
We also observe that when the training distribution average $\bar{p}_{u,i}^n$ is small (e.g., Dist. 1), the adaptation works better for the scenarios with small $p_u^n$. Similarly, when $\bar{p}_{u,i}^n$ is large (e.g., Dist. 5), the adaptation works better around high-vulnerability tasks. 

To evaluate the defense performance, we randomly sample a finite collection of scenarios $\hat{\Theta}_{test}$ i.i.d. from the underlying distribution (different from training sample $\widehat{\Theta}$). For each selected scenario, we use the one-shot adapted policy to evaluate our meta policy. We compare the average cost using the adapted meta policy
$\overline{U}(\tau_{meta})=1/|\hat{\Theta}_{test}|\sum_{\theta}U_{\theta}(\phi(\tau_{meta},\theta))$ with the average cost always using the optimal policy at distribution average $\overline{U}(\tau_{avg}) = 1/|\hat{\Theta}_{test}|\sum_{\theta}U_{\theta}(\tau_{avg})$. It should be noted that there exist testing scenarios that the defender had not encountered during the training process.

Table~\ref{tab:adapt_sys} compares the average cost between the adapted meta policy and distribution average policy. We observe that our meta policy outperforms the distribution-average policy under all $5$ distributions as both the average cost and standard deviation are lower. It indicates that the meta policy has the adaptation ability to the sampled scenarios and provides a lower average cost compared to always using the distribution average policy. SA-ZTD can provide a tailored defense policy against different system configurations.

         {\renewcommand{\arraystretch}{0.9}
        \begin{table}[t]%{width=0.45\textwidth}
            \centering
            \caption{Adaptability with different system vulnerability.}
            \label{tab:adapt_sys}
                \begin{tabular}{|c|c|c|c|}
                \hline
                \rowcolor{Gray}
                \textbf{Distribution} & \textbf{Average $\bar{p}_u^n$} &\textbf{$\overline{U}(\pi_{meta})$} & \textbf{$\overline{U}(\pi_{avg})$}\\
                \hline
                Dist. 1 & 0.05 & 18.98 $\pm$ 1.01 & 19.07 $\pm$ 1.45\\
                \hline
                Dist. 2 & 0.15 & 22.69 $\pm$ 1.50 & 23.63 $\pm$ 1.68\\
                \hline
                Dist. 3 & 0.35 & 29.10 $\pm$ 1.54 & 29.45 $\pm$ 1.77\\
                \hline
                Dist. 4 & 0.55 & 30.63 $\pm$ 1.70 & 31.51 $\pm$ 1.90\\
                \hline
                Dist. 5 & 0.65 & 31.25 $\pm$ 1.23 & 32.01 $\pm$ 1.41 \\
                \hline
                \end{tabular}
                \vskip -2mm
         \end{table}}

     {\renewcommand{\arraystretch}{0.9}
        \begin{table}[t]%{width=0.45\textwidth}
            \centering
            \caption{Adaptability with different attacker's stealthiness.}
            \label{tab:adapt_att}
                \begin{tabular}{|c|c|c|c|}
                \hline
                \rowcolor{Gray}
                \textbf{Distribution} & \textbf{Average $\bar{p}_a^n$} &\textbf{$\overline{U}(\pi_{meta})$} & \textbf{$\overline{U}(\pi_{avg})$}\\
                \hline
                Dist. 1 & 0.65 &  25.46 $\pm$ 0.41 & 26.01 $\pm$ 0.59\\
                \hline
                Dist. 2 & 0.7 & 27.89 $\pm$ 0.54 & 28.40 $\pm$ 0.77\\
                \hline
                Dist. 3 & 0.8 & 29.08 $\pm$ 0.53 & 31.10 $\pm$ 0.55\\
                \hline
                Dist. 4 & 0.9 & 32.98 $\pm$ 0.79 & 33.52 $\pm$ 1.03\\
                \hline
                Dist. 5 & 0.95 & 34.15 $\pm$ 0.74 & 34.23 $\pm$ 0.57  \\
                \hline
                \end{tabular}
                \vskip -6mm
         \end{table}}

\vspace*{-0.4em}
\subsection{Attacker's Stealthiness}
%Another uncertainty in the zero-trust defense scenario is the attacker's capability. 
%We investigate the explainability and adaptability of our model when encountering different attacker types. 
To illustrate the results, we keep the baseline settings and let $p^n_u=0.5$.
The attacker's stealthiness is captured by $p_a^n$ and we consider $\max\{p_u^n,p_u^n-p_u^d+p_a^d\}\leq p_a^n\leq 1$, i.e., $p_a^n\in[0.6,1]$, to keep the threshold policy \cite{ge22trust}. Each $p_a^n$ value represents one scenario $\theta=(p_a^n)$ in the model.

\subsubsection{\textbf{Explainability}}

To evaluate the meta policy under different attacker distributions, again, we consider $5$ generalized Beta distributions $p_i^\prime(\theta,\alpha_i,\beta_i)$ on support $\theta=p_a^n\in[0.6,1]$ with mean values $\bar{p}_{a,i}^n \in\{0.65,0.7,0.8,0.9,0.95\}$, respectively. The meta policy is summarized in Figure~\ref{fig:pap_policy}. 
The single scenario optimal policy indicates that the system must increase the trust threshold $\tau$ if the attacker is more capable.  
As shown in the figure, the adapted policies can capture the relationship between the attacker's stealthiness and defense policy.
When $\bar{p}_a^n$ is closer to $p_{a,min}^n$, the meta policy is closer to $\tau_{min}$. When $\bar{p}_a^n$ is closer to $p_{a,max}^n$, the meta policy is also closer to $\tau_{max}$.
The observations concur with the common explanation that if the attacker is more likely to be stealthy, the defense system needs to guard up and increase the meta-trust threshold. Conversely, if the attacker is less capable, the zero-trust engine could be more tolerant of the agent with a lower trust score.

\subsubsection{\textbf{Adaptability}}

We investigate the adaptation ability of our policy to different attackers' stealthiness levels in Figure~\ref{fig:pap_adapt}. After adaptation, the updated policy successfully adapts to each underlying scenario as it generates a lower trust threshold with small $p_a^n$ and provides a higher threshold with large $p_a^n$.

Finally, we randomly sample a finite set of scenarios $\hat{\Theta}_{test}$ i.i.d. from the underlying distribution and compare $\overline{U}(\tau_{meta})$ with $\overline{U}(\tau_{avg})$. The results in Table~\ref{tab:adapt_att} demonstrate that the meta policy generates a lower cost with less variance compared to the distribution average cost under all $5$ distributions. Our method can quickly adapt to the specific scenario and provide a utility-wise better defense policy.

\vspace*{-0.5em}
\subsection{Real Attack Data}
\begin{figure}[h!]
\vskip -5mm
\abovecaptionskip=0pt
\centerline{\includegraphics[width=0.8\linewidth]{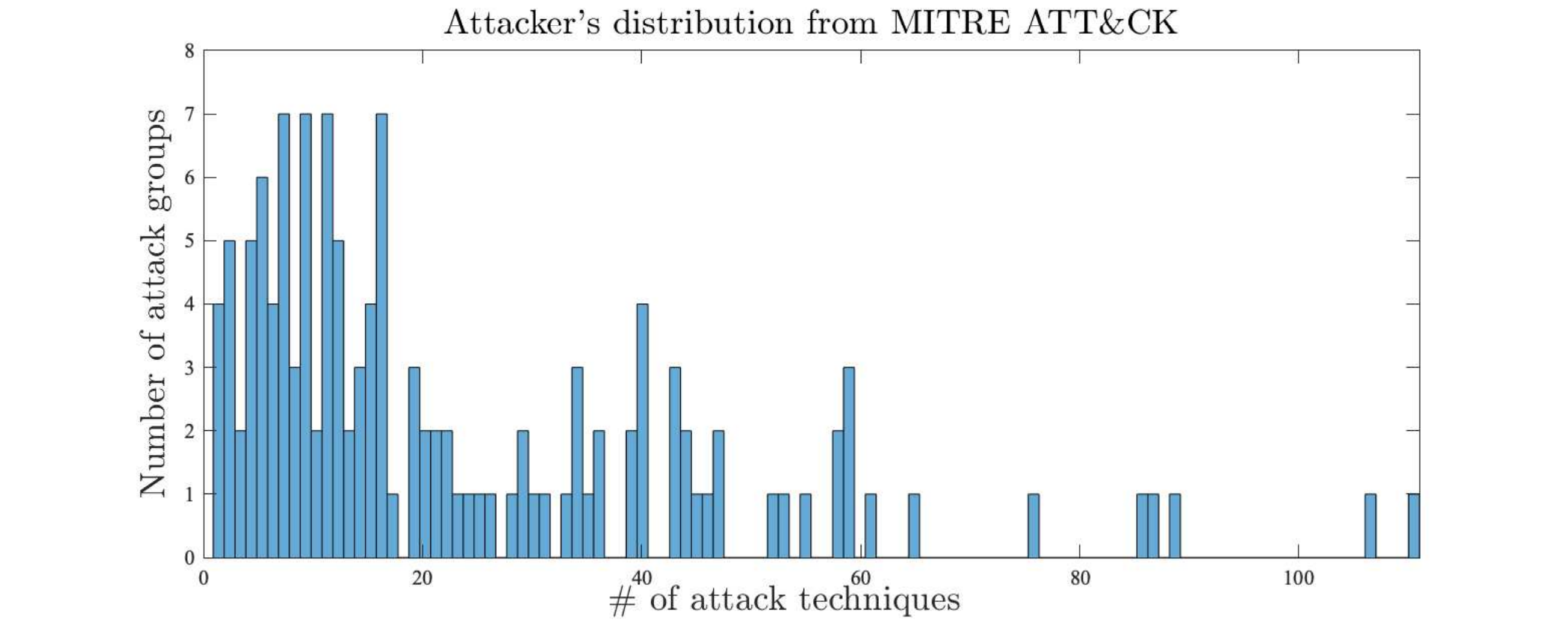}}
  \caption{Attacker's distribution histogram from MITRE ATT\&CK \cite{strom2018mitre}.}
  \vskip -4mm
  \label{fig:real_attack}
\end{figure}
We finally evaluate SA-ZTD on real-world attack distribution. We consider the scenarios $\theta =(p_a^n)$ and collect real-world attack group data from MITRE ATT\&CK \cite{strom2018mitre}. The histogram of the attacker's distribution $\Delta(\Theta)$ is illustrated in Figure~\ref{fig:real_attack}, where the x-axis is the number of attack techniques and the y-axis is the number of attack groups. We use this empirical distribution to approximate the distribution of $p_a^n$ and compute the meta policy.

     {\renewcommand{\arraystretch}{1}
        \begin{table}[h]%{width=0.45\textwidth}
            \centering
            \vskip -2mm
            \caption{System performance with different policies.}
            \label{tab:real}
                \begin{tabular}{|c|c|c|c|}
                \hline
                \rowcolor{Gray}
                \textbf{Distribution} & \textbf{$\overline{U}(\pi_{avg})$}  &\textbf{$\overline{U}(\pi_{meta})$}  & \textbf{$\overline{U}(\pi_{robust})$} \\
                \hline
                Empirical Dist. &28.89 $\pm$ 1.11  &  \textbf{27.89} $\pm$ 0.66 & 32.11 $\pm$ 0.95 \\
                \hline
                Worst-case Dist. & 36.18 $\pm$ 0.86 & 33.75 $\pm$ 1.51 &  \textbf{32.48} $\pm$ 1.39 \\
                \hline
                \end{tabular}
                \vskip -2mm
         \end{table}}

We randomly sample a set of scenarios from empirical and worst-case distributions and compare the average costs with different policies. Table~\ref{tab:real} shows that the meta policy SA-ZTD outperforms the other two cases when we use empirical distribution while the robust policy SR-ZTD has a lower cost when we consider the worst-case distribution.
\vspace*{-0.3em}
\section{Conclusion}
\vskip -1mm
We have formulated a scenario-agnostic zero-trust defense (SA-ZTD) and obtained an explainable and adaptable trust-threshold policy that activates defense based on the agent's trust score. We have developed an algorithm using first-order meta-learning to learn the SA-ZTD policy and extend it to robust defense in response to real-world data. 
Experiments have demonstrated the explainability and adaptability of the proposed model.

\vspace*{-0.7em}
\bibliographystyle{ieeetr}
\bibliography{ref.bib}
\end{document}

%% file: macro_packages_ver10.tex
\usepackage{dsfont}
\usepackage{amsfonts}
\usepackage{amsmath}
\usepackage{amssymb}
\usepackage{amsthm}
\usepackage{pifont}

\DeclareMathOperator*{\argmin}{arg\,min}
\usepackage{booktabs}
\usepackage{multicol}
\usepackage{multirow}
\usepackage{tabularx}
\usepackage{diagbox}
\usepackage{graphicx}
\usepackage{subfigure}
\usepackage{array}
\usepackage{fancyhdr}
\usepackage{hyperref}
 \hypersetup{
     colorlinks=true,
     linkcolor=blue,
     filecolor=blue,
     citecolor = black,      
     urlcolor=cyan,
     }
\usepackage{cleveref}
\usepackage{xcolor}
\usepackage{tikz}
\usepackage{verbatim}
\usepackage{listings}
\definecolor{mygreen}{RGB}{28,172,0} % color values Red, Green, Blue
\definecolor{mylilas}{RGB}{170,55,241}

\usepackage{matlab-prettifier}
\usepackage{color}
\usepackage[shortlabels]{enumitem}

\usepackage{algorithm}
\usepackage[noend]{algpseudocode}
\makeatletter
\def\BState{\State\hskip-\ALG@thistlm}
\makeatother
\newcommand{\E}{\mathbb{E}}
\newcommand{\pr}{\mathbb{P}}

\newcommand{\R}{\mathbb{R}}
\newcommand{\Z}{\mathbb{Z}}

\newtheorem{definition}{Definition}

\usepackage{color, colortbl}
\usepackage{tabularx}
\definecolor{Gray}{gray}{0.9}
\usepackage{enumitem}
\usepackage[font={small,it},skip=3pt]{caption}